\documentclass[11pt,a4paper]{article}
\usepackage[hyperref]{acl2020}
\usepackage{times}
\usepackage{latexsym}

\usepackage{graphicx}
\usepackage{subcaption}
\usepackage{enumitem}
\usepackage{booktabs}
\usepackage{multirow}
\usepackage{makecell}

\aclfinalcopy %

\newcommand{\sa}{\textsc{sa}}
\newcommand{\nli}{\textsc{nli}}
\newcommand{\pp}{\textsc{pp}}
\newcommand{\doc}{\textsc{doc}}
\newcommand{\la}{\textsc{la}}

\newcommand{\fireemoji}{\includegraphics[height=\fontcharht\font`\B]{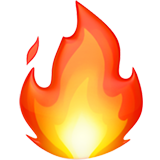}}
\newcommand{\snowflakeemoji}{\includegraphics[height=\fontcharht\font`\B]{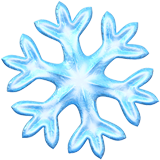}}

\newcommand{\insertTextrayFigure}{
\begin{figure*}[t]
\begin{center}
\begin{subfigure}[b]{0.3\linewidth}
    \centering
    \includegraphics[height=3.5cm]{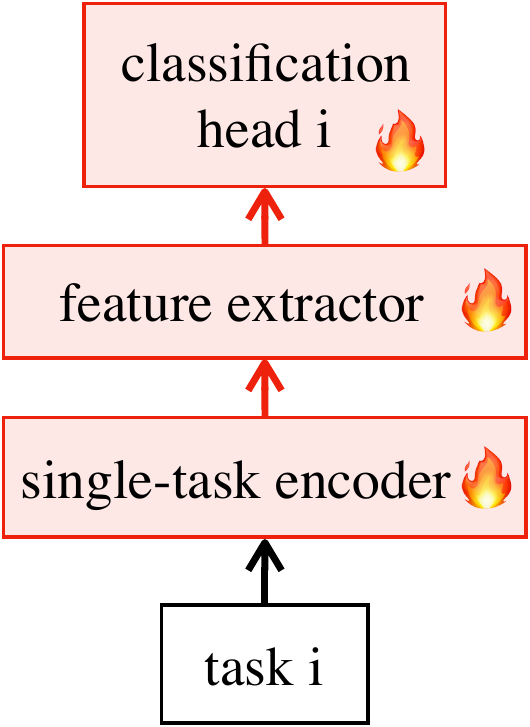}
    \caption{Single-task finetuning.}
    \label{fig:textray:st}
\end{subfigure}
\begin{subfigure}[b]{0.3\linewidth}
    \centering
    \includegraphics[height=3.5cm]{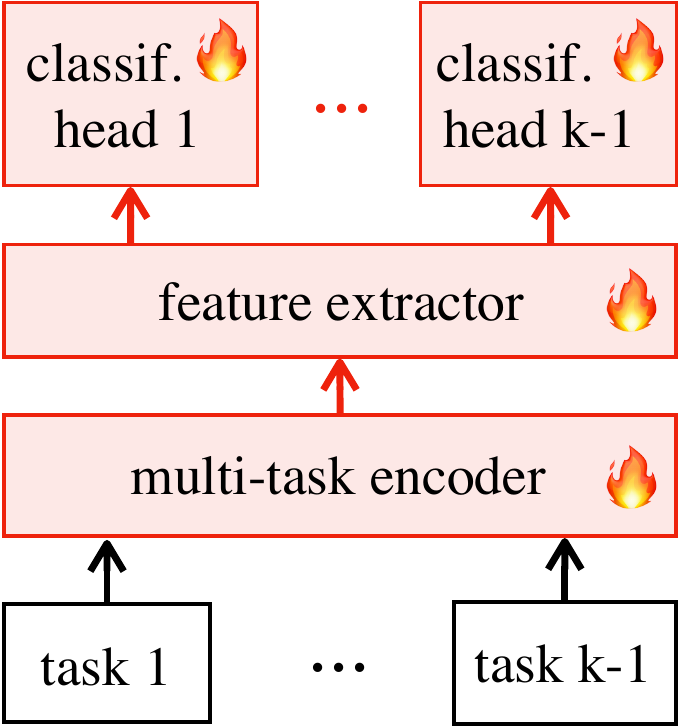}
    \caption{Multi-task pre-training.}
    \label{fig:textray:mt}
\end{subfigure}
\begin{subfigure}[b]{0.34\linewidth}
    \centering
    \includegraphics[height=3.5cm]{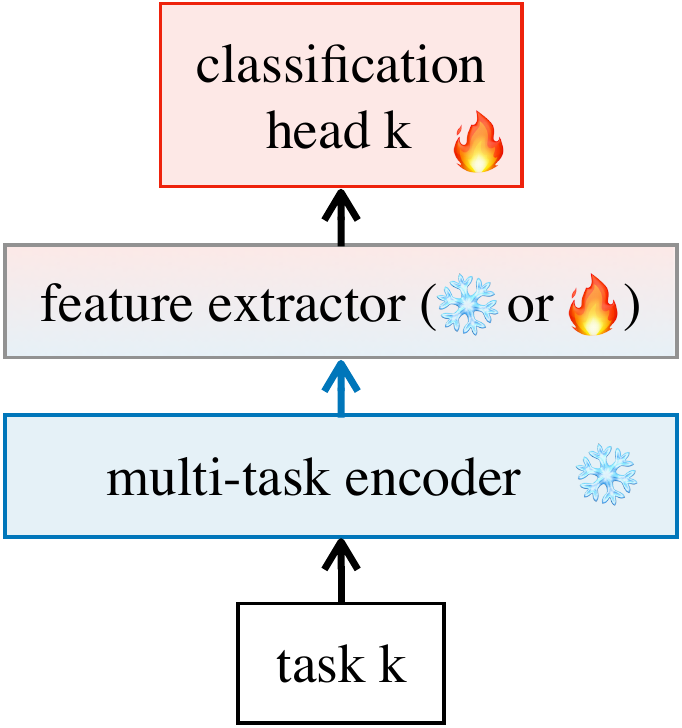}
    \caption{Leave-one-task-out finetuning.}
    \label{fig:textray:loto}
\end{subfigure}
\end{center}
\caption{
An illustration of the finetuning approaches explored in this work.
(a) In \emph{single-task finetuning}, an encoder model is fine-tuned end-to-end for a given task.
(b) In \emph{multi-task pre-training}, an encoder model is jointly trained over $k-1$ tasks, each with their own classification head.
(c) In \emph{leave-one-task-out finetuning}, a multi-task encoder is frozen and used to extract features for an unseen ($k^{th}$) task.
Following \citet{peters2019tune}, we use \fireemoji{} and \snowflakeemoji{} to denote components that are fine-tuned for each task or frozen, respectively.
}
\label{fig:textray}
\end{figure*}
}

\newcommand{\insertMHAFigure}{
\begin{figure}[t]
\begin{center}
\includegraphics[height=3.5cm]{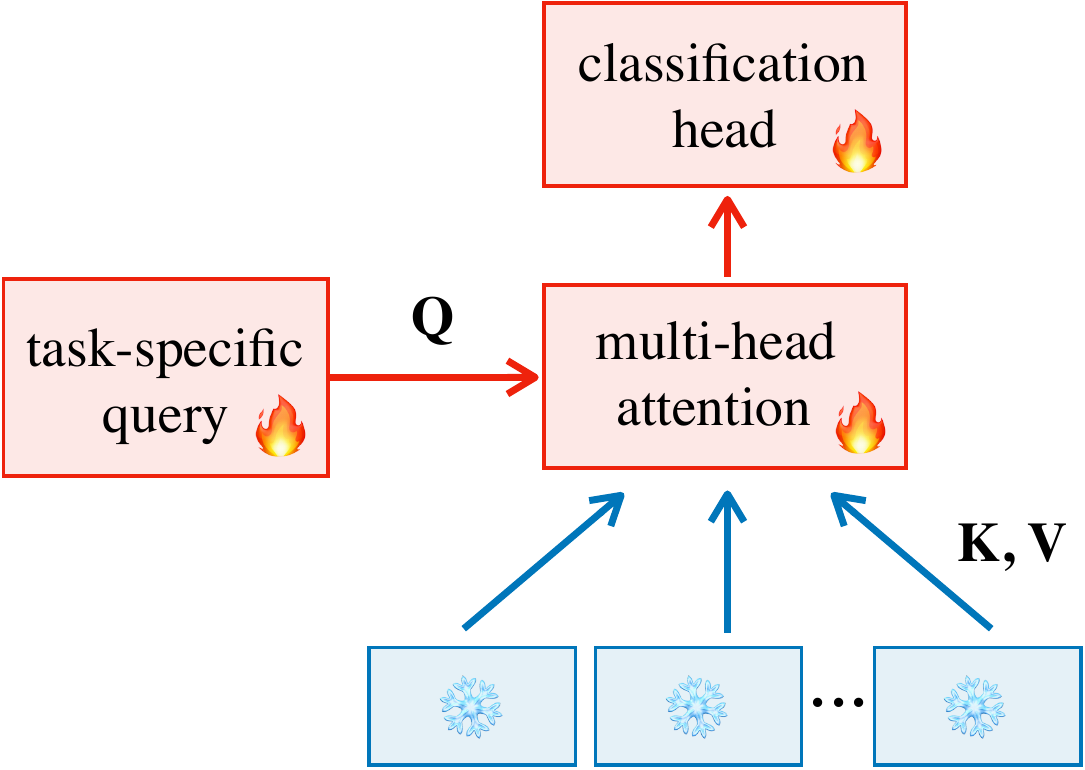}
\end{center}
\caption{
Our proposed multi-head attention pooler.
The extracted features are frozen (\snowflakeemoji{}) and used as both the keys (\textbf{K}) and values (\textbf{V}).
Each task has its own query (\textbf{Q}), multi-head attention module and classification head, all of which are fine-tuned (\fireemoji{}).
}
\label{fig:mha}
\end{figure}
}

\newcommand{\insertPoolerFigure}{
\begin{figure}[t]
\begin{center}
\includegraphics[height=4cm]{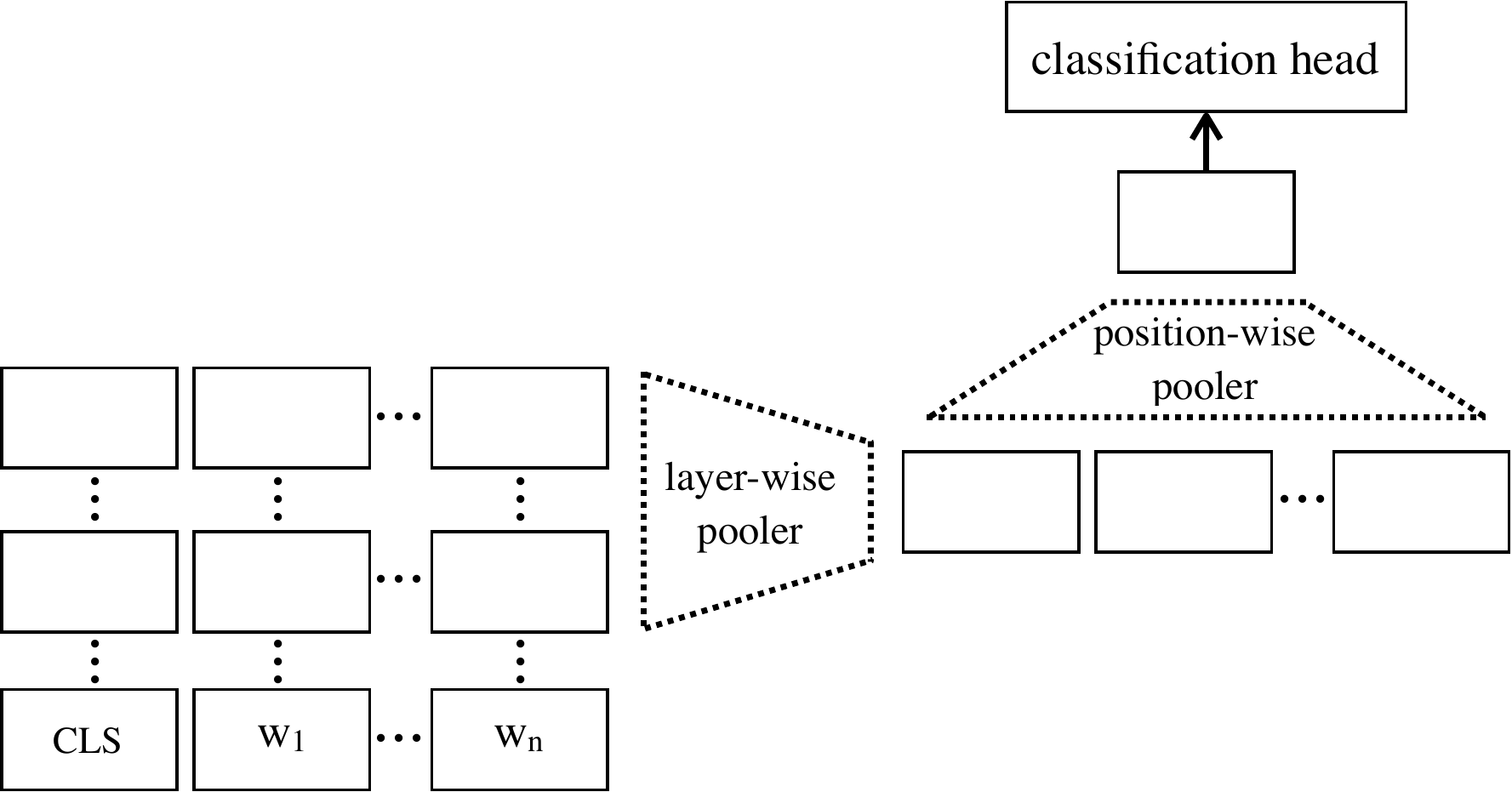}
\end{center}
\caption{
Features are extracted from multiple encoder layers, pooled across layers, then positions, and finally passed to a task-specific classification head.
Some feature extraction and pooling approaches have additional task-specific parameters that require finetuning.
}
\label{fig:pooling}
\end{figure}
}

\newcommand{\insertFLOPsVSNTasksFigure}{
\begin{figure}[t]
\begin{center}
\includegraphics[width=\linewidth]{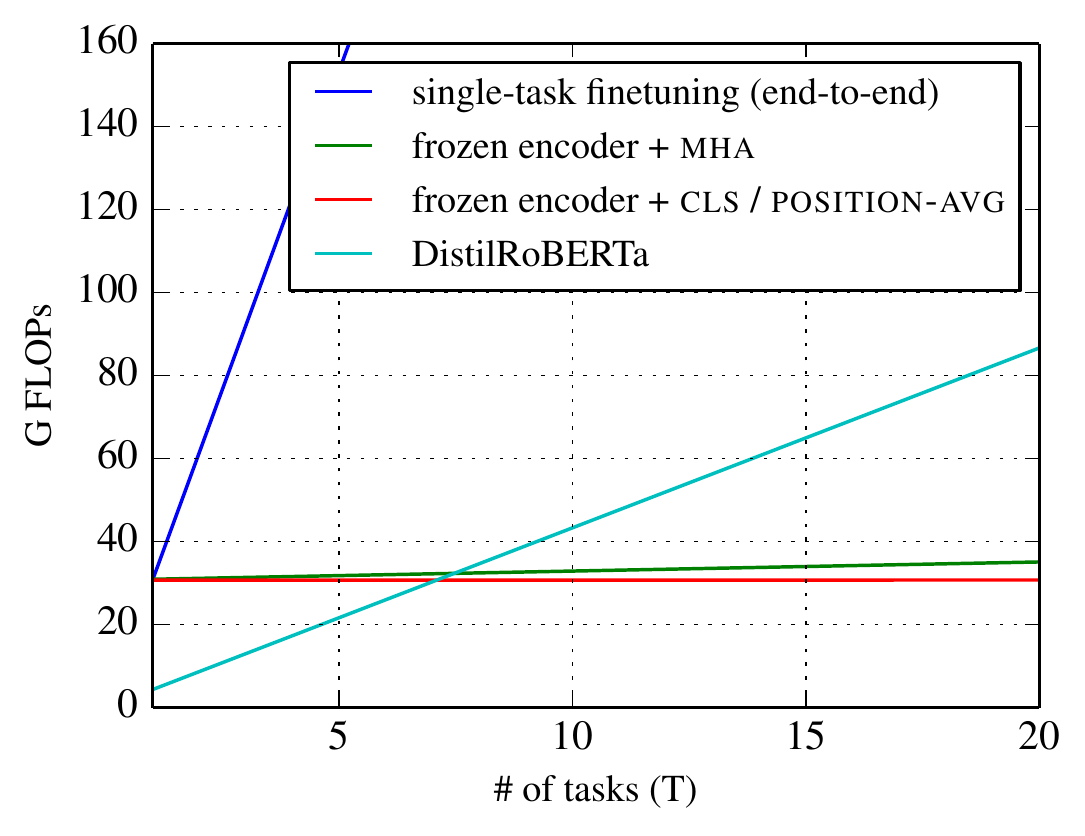}
\end{center}
\caption{Estimated computational cost (in FLOPs) to run RoBERTa inference for $T$ tasks over a single input.
The cost for single-task models grows linearly with the number of tasks, whereas approaches based on a frozen encoder are much more efficient.
Distilled models are particularly efficient when the number of tasks is small, but the cost scales linearly and becomes less efficient than a frozen encoder when the number of tasks $T>7$.}
\label{fig:flops_vs_ntasks}
\end{figure}
}
\newcommand{\insertTasksTable}{
\begin{table}[t]
\scriptsize
\begin{center}
\resizebox{1\linewidth}{!}{
\begin{tabular}[b]{llccc}
\toprule
\bf task & \bf type & \bf \# train & \bf \# dev & \bf \# label \\
\midrule
MNLI & \nli{} & 393K & 20K & 3 \\
QNLI & \nli{} & 105K & 5.4K & 2 \\
QQP & \pp{} & 364K & 391K & 2 \\
RTE & \nli{} & 2.5K & 3K & 2 \\
SST-2 & \sa{} & 17K & 1.8K & 2 \\
MRPC & \pp{} & 3.7K & 1.7K & 2 \\
CoLA & \la{} & 8.5K & 1K & 2 \\
AG-news & \doc{} & 120K & 7.6K & 4 \\
Amazon-5 & \sa{} & 3M & 650K & 5 \\
Amazon-2 & \sa{} & 3.6M & 400K & 2 \\
Yelp-5 & \sa{} & 650K & 50K & 5 \\
Yelp-2 & \sa{} & 560K & 38K & 2 \\
DBpedia & \doc{} & 560K & 70K & 14  \\
\bottomrule
\end{tabular}
}
\caption{Task statistics.\label{tab:tasks}}
\end{center}
\vspace{-0.4cm}
\end{table}
}

\newcommand{\insertMainTable}{
\begin{table*}[p]
\begin{center}
\begin{tabular}[b]{p{5.5cm}ccccccc}
\toprule
\bf model & \bf G FLOPs & \bf \sa{} & \bf \nli{} & \bf \pp{} & \bf \doc{} & \bf \la{} & \bf \textsc{average} \\
\midrule
\addlinespace
\multicolumn{8}{l}{(a) \textit{Single-task finetuning (end-to-end):}} \\
\addlinespace
BERT & - & 86.8 & 83.1 & 89.7 & 97.1 & 83.1$^{*}$ & 87.6 \\
XLNet & - & 87.6 & 89.2 & 90.5 & \bf 97.4 & 84.5$^{*}$ & 89.5 \\
DistilRoBERTa & 61 & 86.6 & 80.7 & 89.6 & 97.1 & 84.3 & 87.1 \\
RoBERTa & 430 & 88.2 & 91.3 & 91.8 & \bf 97.4 & 86.3 & 90.5 \\
\ + leave-one-task-out multi-task & 430 & \bf 88.2 & \bf 91.6 & \bf 92.1 & \bf 97.4 & \bf 87.2 & \bf 90.7 \\
\quad pre-training \\
\addlinespace
\midrule
\addlinespace
\multicolumn{8}{l}{(b) \textit{Single-task finetuning (frozen encoder):}} \\
\addlinespace
RoBERTa \\
\ + \textsc{last-layer / cls} & 31 & 80.8 & 58.8 & 68.2 & 94.9 & 69.1 & 75.5 \\
\ + \textsc{last-layer / position-avg} & 31 & 80.3 & 63.0 & 75.6 & 95.0 & 75.0 & 77.7 \\
\ + \textsc{last-layer / mha} & 34 & 86.1 & 72.7 & 79.0 & 96.7 & 80.2 & 83.3 \\
\ + \textsc{layer-avg / mha} & 34 & 86.9 & 77.7 & \bf 83.0 & \bf 96.9 & \bf 82.7 & 85.5 \\
\ + \textsc{learned-comb / mha} & 34 & \bf 87.0 & \bf 78.0 & 82.8 & \bf 96.9 & 82.5 & \bf 85.6 \\
\addlinespace
\midrule
\addlinespace
\multicolumn{8}{l}{(c) \textit{Leave-one-task-out finetuning (frozen multi-task encoder):}} \\
\addlinespace
RoBERTa \\
\ + \textsc{last-layer / cls} & 31 & 87.4 & 82.8 & 81.8 & 94.9 & 76.4 & 85.9 \\
\ + \textsc{last-layer / position-avg} & 31 & 87.4 & 83.0 & 81.9 & 95.1 & 77.1 & 86.0 \\
\ + \textsc{last-layer / mha} & 34 & 87.5 & 84.6 & 83.5 & 96.2 & 77.2 & 86.8 \\
\ + \textsc{layer-avg / mha} & 34 & \bf 87.9 & 87.8 & \bf 85.7 & 96.8 & 82.4 & 88.4 \\
\ + \textsc{learned-comb / mha} & 34 & \bf 87.9 & \bf 87.9 & \bf 85.7 & \bf 96.9 & 82.3 & \bf 88.5 \\
\addlinespace
RoBERTa (8-bit quantization) \\
\ + \textsc{layer-avg / mha} & 34 & \bf 87.9 & 87.7 & \bf 85.7 & 96.8 & \bf 82.6 & 88.4 \\
\addlinespace
RoBERTa (1-bit quantization) \\
\ + \textsc{layer-avg / mha} & 34 & 87.8 & 87.1 & 84.6 & 96.6 & 81.3 & 88.0 \\
\addlinespace
\midrule
\addlinespace
\multicolumn{8}{l}{(d) \textit{Leave-one-task-group-out finetuning (frozen multi-task encoder):}} \\
\addlinespace
RoBERTa \\
\ + \textsc{layer-avg / mha} & 31 & 87.0 & 81.3 & 85.3 & 96.7 & 82.4 & 86.6 \\
\addlinespace
\midrule
\addlinespace
\multicolumn{8}{l}{(e) \textit{Multi-task pre-training over all tasks (frozen multi-task encoder; no additional finetuning):}} \\
\addlinespace
RoBERTa \\
\ + \textsc{last-layer / cls} & 31 & 87.7 & 89.6 & 89.3 & 97.2 & 82.6 & 89.3 \\
\addlinespace
\bottomrule
\end{tabular}
\caption{Results on 14 tasks, grouped by task type (see Section~\ref{sec:tasks}).
We consider different layer-wise and position-wise pooling strategies introduced in Section~\ref{sec:pooling}.
We also report the estimated inference cost for 14 tasks (in G FLOPs) for each strategy.
Bold results indicate the most accurate method in each section.
BERT results are from \citet{yang2019xlnet} and \citet{sun2019fine}.
XLNet results are from \citet{yang2019xlnet}.
DistilRoBERTa and RoBERTa results are recomputed ourselves.
Full results for each task is given in the Appendix.
(*) we recomputed accuracy for CoLA, since BERT and XLNet originally reported a different metric.
\label{tab:all}}
\end{center}
\end{table*}
}

\newcommand{\insertExtendedTable}{
\begin{table*}
\begin{center}
\resizebox{\linewidth}{!}{
\begin{tabular}[b]{lccccccccccccccc}
\toprule
Model & MNLI & QNLI & QQP & RTE & SST2 & MRPC & CoLA & IMDB & AG & Amzn5 & Amzn2 & Yelp5 & Yelp2 & DBpd & Avg \\
\midrule
\addlinespace
\multicolumn{16}{l}{(a) \textit{Single-task finetuning (end-to-end):}} \\
\addlinespace
BERT & 86.6 & 92.3 & 91.3 & 70.4 & 93.2 & 88.0 & 83.1 & 95.5 & 94.8 & 65.8 & 97.4 & 70.7 & 98.1 & 99.4 & 87.6\\
XLNet & 89.8 & 93.9 & 91.8 & 83.8 & 95.6 & 89.2 & 84.5 & 96.2 & 95.5 & 67.7 & 97.4 & 70.7 & 98.1 & 99.4 & 89.5\\
DistilRoBERTa & 83.9 & 91.0 & 91.2 & 67.2 & 93.5 & 88.0 & 84.3 & 94.4 & 94.9 & 66.3 & 97.1 & 70.5 & 97.9 & 99.3 & 87.1\\
RoBERTa & 90.3 & 94.6 & 92.3 & 88.9 & 96.7 & 91.3 & 86.3 & 96.4 & 95.4 & 67.9 & 97.6 & 71.9 & 98.4 & 99.3 & 90.5\\
\ + leave-one-task-out multi-task & 90.3 & 94.6 & 92.2 & 89.9 & 96.7 & 92.1 & 87.2 & 96.6 & 95.5 & 68.0 & 97.6 & 72.2 & 98.4 & 99.3 & 90.7\\
\quad pre-training \\
\addlinespace
\midrule
\addlinespace
\multicolumn{16}{l}{(b) \textit{Single-task finetuning (frozen encoder):}} \\
\addlinespace
RoBERTa \\
\ + \textsc{last-layer / cls} & 55.7 & 67.4 & 67.4 & 53.3 & 84.0 & 69.0 & 69.1 & 89.1 & 91.0 & 58.2 & 94.6 & 62.7 & 96.2 & 98.7 & 75.5 \\
\ + \textsc{last-layer / position-avg} & 56.7 & 72.7 & 80.3 & 59.7 & 88.1 & 70.9 & 75.0 & 87.2 & 91.5 & 57.3 & 93.7 & 60.6 & 95.0 & 98.4 & 77.7 \\
\ + \textsc{last-layer / mha} & 75.7 & 81.6 & 86.0 & 60.7 & 92.5 & 71.9 & 80.3 & 94.3 & 94.1 & 65.1 & 96.9 & 69.9 & 98.0 & 99.3 & 83.3 \\
\ + \textsc{layer-avg / mha} & 83.1 & 87.3 & 88.1 & 62.8 & 94.3 & 77.9 & 82.7 & 95.5 & 94.4 & 65.9 & 97.2 & 70.6 & 98.2 & 99.3 & 85.5\\
\ + \textsc{learned-comb / mha} & 83.4 & 87.4 & 88.1 & 63.1 & 94.4 & 77.4 & 82.5 & 95.5 & 94.5 & 66.0 & 97.2 & 70.8 & 98.2 & 99.3 & 85.6 \\
\addlinespace
\midrule
\addlinespace
\multicolumn{16}{l}{(c) \textit{Leave-one-task-out finetuning (frozen multi-task encoder):}} \\
\addlinespace
RoBERTa \\
\ + \textsc{last-layer / cls} & 76.2 & 84.3 & 84.7 & 87.8 & 94.0 & 78.8 & 76.4 & 96.7 & 90.7 & 66.4 & 97.6 & 71.2 & 98.7 & 99.1 & 85.9 \\
\ + \textsc{last-layer / position-avg} & 76.3 & 84.6 & 84.7 & 88.2 & 93.8 & 79.1 & 77.1 & 96.7 & 91.1 & 66.3 & 97.6 & 71.0 & 98.7 & 99.0 & 86.0\\
\ + \textsc{last-layer / mha} & 79.8 & 87.6 & 86.2 & 86.5 & 94.1 & 80.8 & 77.2 & 96.7 & 93.2 & 66.5 & 97.6 & 71.5 & 98.7 & 99.2 & 86.8\\
\ + \textsc{layer-avg / mha} & 86.6 & 91.6 & 88.7 & 85.1 & 96.0 & 82.7 & 82.4 & 96.7 & 94.4 & 66.8 & 97.6 & 71.7 & 98.7 & 99.3 & 88.4\\
\ + \textsc{learned-comb / mha} & 87.0 & 91.7 & 88.8 & 85.1 & 96.1 & 82.7 & 82.3 & 96.7 & 94.4 & 66.8 & 97.6 & 71.8 & 98.7 & 99.3 & 88.5\\
\addlinespace
RoBERTa (8-bit quantization) \\
\ + \textsc{layer-avg / mha} & 86.7	& 91.5 & 88.6 & 85.1 & 96.0 & 82.7 & 82.6 & 96.7 & 94.4 & 66.7 & 97.6 & 71.6 & 98.7 & 99.3 & 88.4 \\
\addlinespace
RoBERTa (1-bit quantization) \\
\ + \textsc{layer-avg / mha} & 85.5 & 90.8 & 88.2 & 85.1 & 95.8 & 81.0 & 81.3 & 96.5 & 93.9 & 66.6 & 97.6 & 71.4 & 98.7 & 99.3 & 88.0\\
\addlinespace
\midrule
\addlinespace
\multicolumn{16}{l}{(d) \textit{Leave-one-task-group-out finetuning (frozen multi-task encoder):}} \\
\addlinespace
RoBERTa \\
\ + \textsc{layer-avg / mha} & 85.2 & 90.4 & 88.7 & 68.3 & 94.3 & 82.0 & 82.4 & 95.6 & 94.2 & 65.8 & 97.2 & 70.6 & 98.2 & 99.3 & 86.6 \\
\addlinespace
\midrule
\addlinespace
\multicolumn{16}{l}{(e) \textit{Multi-task pre-training over all tasks (frozen multi-task encoder; no additional finetuning):}} \\
\addlinespace
RoBERTa \\
\ + \textsc{last-layer / cls} & 89.3 & 93.7 & 89.6 & 85.9 & 94.8 & 89.0 & 82.6 & 96.7 & 95.1 & 66.5 & 97.5 & 71.9 & 98.6 & 99.3 & 89.3 \\
\addlinespace
\bottomrule
\end{tabular}
}
\caption{Extended results table for all 14 tasks. See Table~\ref{tab:all} for more details.
\label{tab:extended}}
\end{center}
\end{table*}
}

\newcommand{\insertStorageTable}{
\begin{table}[t]
\begin{center}
\resizebox{1\linewidth}{!}{
\begin{tabular}[b]{ccccc}
\toprule
\bf layer-wise & \bf position-wise & \multirow{2}{*}{\bf fp16} & \multirow{2}{*}{\bf int8} & \multirow{2}{*}{\bf bool} \\
\bf pooling & \bf pooling \\
\midrule
\multirowcell{3}{\rule{0pt}{2.5ex} \textsc{last-layer}\\ or\\ \textsc{layer-avg}} & \textsc{cls} or & \multirow{2}{*}{2K} & \multirow{2}{*}{1K} & \multirow{2}{*}{128} \\
& \textsc{position-avg} & \\
\cmidrule{2-5}
& \textsc{mha} & 100K & 50K & 6K \\
\midrule
\multirowcell{1}{\rule{0pt}{2ex}\textsc{learned-comb}} & \textsc{mha} & 2.3M & 1.2M & 150K \\
\bottomrule
\end{tabular}
}
\caption{Estimated storage cost (in bytes) to store features for a 50 token input for various pooling, quantization methods.
\label{tab:storage}}
\end{center}
\vspace{-0.4cm}
\end{table}
}

\title{General Purpose Text Embeddings from Pre-trained Language Models for Scalable Inference}

\author{Jingfei Du\thanks{\ \ Equal contribution.} \quad
  Myle Ott\footnotemark[1] \quad
  \bf Haoran Li \quad 
  Xing Zhou \quad
  Veselin Stoyanov \\
  Facebook AI
   \\
   \texttt{\{jingfeidu,myleott,haoranli,xingz,ves\}@fb.com}
  }

\date{}

\begin{document}
\maketitle
\begin{abstract}
The state of the art on many NLP tasks is currently achieved by large pre-trained language models, which require a considerable amount of computation.
We explore a setting where many different predictions are made on a single piece of text.
In that case, some of the computational cost during inference can be amortized over the different tasks using a shared text encoder.
We compare approaches for training such an encoder and show that encoders pre-trained over multiple tasks generalize well to unseen tasks.
We also compare ways of extracting fixed- and limited-size representations from this encoder, including different ways of pooling features extracted from multiple layers or positions.
Our best approach compares favorably to knowledge distillation, achieving higher accuracy and lower computational cost once the system is handling around 7 tasks.
Further, we show that through binary quantization, we can reduce the size of the extracted representations by a factor of 16 making it feasible to store them for later use.
The resulting method offers a compelling solution for using large-scale pre-trained models at a fraction of the computational cost when multiple tasks are performed on the same text.
\end{abstract}

\section{Introduction} \label{sec:intro}

Large pre-trained language models achieve state-of-the-art performance on many Natural Language Processing (NLP) tasks~\cite{peters2018deep,radford2018gpt,devlin2018bert}.
However, inference for these models requires significant computational resources, which limits their practical use.
Recent trends show that scaling models up~\cite{liu2019roberta,lan2019albert,raffel2019exploring,li2020train} in terms of computation still improves end task performance, raising questions about whether and how the most accurate models can be applied in real-world settings.

This computational burden is further exacerbated by the need to fine-tune end-to-end a separate model for each task.
Since each model has a new set of parameters, none of the computation can be shared by models for different tasks during inference.
This is particularly inefficient in real-world settings that require making multiple predictions about each input.
For example, given a news article, we may want to predict its topic~\cite{zhang2015character}, sentiment~\cite{pang2004sentimental,maas-etal-2011-learning,socher2013recursive,zhang2015character}, overall text quality~\cite{pitler2008revisiting}, whether it is humorous~\cite{yang2015humor} or offensive~\cite{schmidt2017survey,zampieri2019semeval} and so on.

Knowledge Distillation (KD) is one way of reducing the computation required by large pre-trained LMs~\cite{hinton2015distilling,sanh2019distilbert}. However, there is a sizeable gap in accuracy between the best models using knowledge distillation and the full fine-tuned models. Another way of speeding up computation is through system optimizations such as quantization and operator fusion~\cite{zafrir2019q8bert}. These techniques can reduce the amount of computation significantly, but may not be sufficient by themselves and can be combined with the methods we discuss.    

\insertTextrayFigure{}

In this paper we look at new ways to make inference computationally efficient focusing on the case where different models (models for different tasks) are run over the same piece of text. We propose new methods to run multiple task-specific models in a way that amortizes the computation over the different tasks. The central idea is to compute the activations for the full model once and use smaller task-specific models on top of it. We explore three possible ways for sharing computation.

The first solution is inspired by work on general purpose text encoders~\cite{kiros2015skip,hill-etal-2016-learning-distributed,conneau2017infersent,subramanian2018learning}, which produce fixed-size representations (i.e., sentence embeddings) that can be shared across tasks.
We add only small task-specific layers on top of these fixed-size representations so that the computational cost is dominated by the encoder, which is amortized over tasks.
Unfortunately, when evaluated on unseen tasks, we find that models that rely on fixed-size representations often underperform single-task baselines by a large margin, in agreement with past work~\cite{subramanian2018learning,peters2019tune,raffel2019exploring,wang2019can}.

The second solution is a \emph{multi-task system}~\cite{caruana1997multitask,collobert2008unified,ruder2017overview}, where a single model is jointly trained to handle many tasks (see Figure~\ref{fig:textray:mt}).
If most layers are shared, the overall inference cost can be nearly $k$ times less than for $k$ separate single-task models, while providing competitive task accuracy~\cite{liu2019mtdnn,raffel2019exploring,wang2019can}.
However, multi-task systems work best when the set of tasks is known in advance, since adding new tasks typically requires retraining the multi-task model and reincurring inference costs, thus limiting the usefulness of this approach in real-world systems where new classification tasks may be introduced periodically.

We propose a third solution: a multi-task encoder that is shared across tasks and produces \emph{limited-size} representation that grow with the length of the input, similar to contextualized word representations~\cite{peters2018deep}.
We evaluate our representations on 14 text classification tasks using a \emph{leave-one-task-out} evaluation protocol (see Figure~\ref{fig:textray:loto}), where a multi-task encoder model is trained on $k-1$ tasks, frozen and used as a static feature extractor for an unseen $k^{th}$ task.\footnote{We consider a \emph{task} to be synonymous with a \emph{dataset}.}
We find an important ingredient to performing well on an unseen ($k^{th}$) task is to extract features from multiple layers and positions of the encoder.
Ultimately, our general purpose encoders offer a better tradeoff between task accuracy and inference cost than either fixed-size representations or distilled models, while requiring minimal additional inference cost to handle new tasks.

We also consider the case in which not all of the predictions can be done at the same time and intermediate representations have to saved. In that context, we study the relationship between representation size and end-task performance.
We find that features extracted by our encoders are amenable to heavy quantization enabling a 16x reduction in the size of the extracted features with negligible impact on unseen task performance.

\section{Related Work} \label{sec:rel_work}

\textbf{Self-supervised pre-training}, typically through language modeling, has advanced the state of the art for many NLP tasks~\cite{peters2018deep,radford2018gpt,devlin2018bert}.
There are two dominant ways of adapting pre-trained models to downstream tasks: (1) finetuning, which often results in the best accuracy~\cite{devlin2018bert}; and (2) feature extraction, which can be significantly more efficient during inference when there are multiple end tasks.
\citet{peters2019tune} compare these and find finetuning outperforms feature extraction for BERT; however, they use features immediately after pre-training, whereas we also consider features after multi-task finetuning.

\textbf{Multi-task learning (MTL)} has a rich history in machine learning~\cite{caruana1997multitask,ruder2017overview} and NLP~\cite{collobert2008unified,luong2016multi}.
Multi-task models can potentially leverage similarities across tasks to achieve higher end-task accuracy than single-task models~\cite{clark2019bam,liu2019mtdnn,phang2018stilts,wang2019can}.
Compared to single-task models, a multi-task model can also be more efficient during inference by sharing computation across tasks.
Most work in multi-task learning assumes that the set of end-tasks is fixed and known in advance and training is performed for all tasks together. This set-up can present challenges in the real world where tasks may require different retraining schedules and new tasks may be frequently added or removed.

\textbf{General purpose text encoders} are usually pre-trained with a mix of supervised and self-supervised training objectives and produce fixed-size representations~\cite{kiros2015skip,hill-etal-2016-learning-distributed,conneau2017infersent,subramanian2018learning}.
Unlike multi-task learning, general purpose text encoders are typically evaluated on \emph{unseen tasks}, which is more representative of real-world settings in which new tasks may be added periodically.
Unfortunately, these approaches often underperform single-task baselines~\cite{mccann2018decanlp,liu2019mtdnn,wang2019can}.

Another line of work has explored adapting pre-trained models by adding additional task-specific capacity at each layer~\cite{houlsby2019parameter}, however these methods do not improve inference efficiency since there is no task-independent computation that can be shared across tasks.

\textbf{Knowledge Distillation}~\cite{bucilua2006model,hinton2015distilling} is a compression technique where a more efficient student model is trained to mimic the behaviour of a larger or ensembled teacher model. A knowledge distilled version of BERT~\cite{sanh2019distilbert} has been proposed to reduce the computation required by these large pre-trained language models. However there is still a sizeable gap in terms of accuracy where DistilRoBERTa reaches 95\% of RoBERTa-base's performance on GLUE while being twice faster.

\textbf{Quantization} and other compression techniques have been explored for word embeddings~\cite{shu2017compressing,tissier2019near} and sentence embeddings~\cite{shen2019learning}.
Recent work has also explored quantization for contextualized word representations, generally showing that quantization-aware training is necessary to achieve reasonable end task performance~\cite{zafrir2019q8bert,fan2020training}.
Quantization is complementary to the approaches we consider and is explored more in Section~\ref{sec:quantization}.

\section{Experimental Setup} \label{sec:exp_setup}

Our goal is to develop text encoders that produce representations which achieve high accuracy for multiple task with little task-specific processing. We first introduce our tasks, encoder models and finetuning framework.

\subsection{Tasks} \label{sec:tasks}

\insertTasksTable{}

We consider 14 text classification tasks, spanning sentiment analysis (\sa{}), natural language inference (\nli{}), paraphrase identification (\pp{}), document categorization (\doc{}) and linguistic acceptability (\la{}).
Tasks are chosen for their diversity and usage in recent related work, ensuring that our baselines are representative of the state of the art.

Full details about each task is given in Table~\ref{tab:tasks}.
The \sa{}, \doc{} and \la{} tasks require making predictions about a single text input, while \nli{} and \pp{} tasks require classifying a pair of text inputs.
For pair tasks we concatenate the text with a special separator token following \citet{liu2019roberta}.
Since many of our tasks are part of evaluation benchmarks such as GLUE~\cite{wang2019glue} and the test sets are not publicly available, we report accuracy on the corresponding development sets.

\subsection{Encoder models} \label{sec:models}

Our encoder models are based on RoBERTa~\cite{liu2019roberta}, an optimized version of BERT~\cite{devlin2018bert} that achieves competitive performance on most of the tasks considered in this work.
We primarily use the public RoBERTa$_{\textsc{large}}$ model consisting of 24 Transformer layers~\cite{vaswani2017attention}, 1024 dimensional representations and 355M parameters.
We refer the reader to \citet{devlin2018bert} for more details about the BERT architecture and \citet{liu2019roberta} for more details about RoBERTa.

We also consider a Knowledge Distilled (KD) version of RoBERTa called DistilRoBERTa~\cite{sanh2019distilbert}, which consists of 6 Transformer layers, 768-dim representations and 82M parameters.
The distilled model contains 1/4 as many parameters and requires 1/7 as much computation (FLOPs) as the full model.
We present a more detailed comparison of the computational requirements for these encoder models in Section~\ref{sec:res:flops}.

\subsection{Fine-tuning} \label{sec:finetuning}

We consider two methods for finetuning encoder models, illustrated in Figure~\ref{fig:textray}.

\subsubsection{Single-task finetuning} \label{sec:single_task}

Single-task finetuning is the most common way of adapting pre-trained language models to a given task (see Figure~\ref{fig:textray:st}).
When applied to large pre-trained models (e.g., RoBERTa) single-task finetuning often results in the best end-task accuracy, but requires the full model to be run for every task and thus has the highest inference costs for a set of $k$ tasks.
Computation can be reduced by using a smaller pre-trained models---including knowledge distilled models (e.g., DistilRoBERTa).

Single-task finetuning serves as our baseline finetuning method.
Our goal is to achieve similar accuracy as large single-task models with reduced inference costs.

\subsubsection{Leave-one-task-out finetuning} \label{sec:loto}

We also consider \emph{leave-one-task-out} finetuning, illustrated in Figures~\ref{fig:textray:mt} and~\ref{fig:textray:loto}.
A multi-task encoder is pre-trained on $k-1$ tasks, then frozen and used as a feature extractor for a $k^{th}$ task.
Freezing the encoder allows us to amortize the inference cost over all tasks.
The leave-one-task-out setup allows us to evaluate generalization on tasks \emph{unseen} in the training of the encoder.
This replicates the real-world setting of adding new tasks to an existing frozen encoder.
Leave-one-task-out finetuning has two stages:

\noindent \textbf{1. Multi-task pre-training:}
We train a single model end-to-end over $k-1$ tasks (Figure~\ref{fig:textray:mt}).
The majority of the encoder weights are shared across tasks, except for a classification head (see Section~\ref{sec:heads}) that is unique to each task.

It is important for the multi-task model to properly weight the training data for different tasks, so that larger tasks do not dominate smaller ones~\cite{raffel2019exploring,wang2019can}.
We adopt a loss-reweighting technique inspired by \citet{raffel2019exploring}.
At each step, we sample a batch of data for every task and update our model according to a weighted sum of the losses.
Each task's loss is weighted according to: $\alpha_i = D_i^{\left(\frac{1}{T}\right)} / \sum_j D_j^{\left(\frac{1}{T}\right)}$, where $D_i$ is the number of training examples for task $i$ and $T$ is a temperature controlling the uniformity of the weights.
When $T=1$, task weights are proportional to data size, and as $T \rightarrow 0$, task weights become more uniform.
We use a fixed temperature of $T=0.1$, which performed best in early experiments.

\noindent \textbf{2. Leave-one-task-out finetuning:}
In the second stage, we freeze the multi-task encoder's weights and use it as a feature extractor for an unseen $k^{th}$ task (see Figure~\ref{fig:textray:loto}).
The extracted features are fed to a new, randomly initialized classification head, which is fine-tuned over the training data for the $k^{th}$ task.
We repeat this process $k$ times, with each task held out once, and report the corresponding held-out task performance.

\subsection{Classification heads} \label{sec:heads}

Each task has a \emph{classification head} that takes the pooled features as input and makes a prediction.
While related work incorporates custom task-specific classification layers~\cite{peters2018deep,peters2019tune,liu2019mtdnn}, we adopt a unified architecture for all tasks.
We follow the original BERT setup~\cite{devlin2018bert} and use a two-layer Multi-Layer Perceptron (MLP) with inner dimension equal to the pooled feature dimension and a \texttt{tanh} activation function.
The classification head is always fine-tuned for the end task.

\section{Feature extraction and pooling} \label{sec:pooling}

\insertPoolerFigure{}

The most common way of extracting features from BERT-like models is by taking the representation of the last Transformer layer corresponding to the special \texttt{CLS} token prepended to the input text sequence~\cite{devlin2018bert}.
More recent work has also explored extracting features from every position and layer, and then linearly combining the layers with task-specific weights~\cite{peters2019tune,tenney2019bert}.

We propose a more general framework for extracting features, shown in Figure~\ref{fig:pooling}.
We extract features from several layers of the encoder and then \emph{pool} them, first across layers and then across positions, before feeding them to a task-specific classification head.
This framework subsumes both the \texttt{CLS} token and weighted layer combination approaches.
We consider several ways of \emph{layer-wise pooling} and \emph{position-wise pooling}:

\paragraph{Layer-wise pooling approaches:}
\begin{itemize}[leftmargin=*]
\setlength\itemsep{0em}
\item \textsc{last-layer}: only use the last layer. This setting is used by \citet{devlin2018bert}.
\item \textsc{layer-avg}: average the last $m$ layers. We tune $m$ for each setting, but find that $m=16$ works best in most cases.
\item \textsc{learned-comb}: learn a task-specific weighted combination over all layers. This setting is used by \citet{peters2019tune} and \citet{tenney2019bert}.
\end{itemize}

\insertMHAFigure{}

\noindent \textbf{Position-wise pooling approaches:}
\begin{itemize}[leftmargin=*]
\setlength\itemsep{0em}
\item \textsc{cls}: extract features from the first position. This setting is used by \citet{devlin2018bert}.
\item \textsc{position-avg}: average features across positions.
\item \textsc{mha}: pool features with a task-specific Multi-Head Attention (MHA) layer~\cite{devlin2018bert}. We learn a task-specific query and use features as the keys and values (see Figure~\ref{fig:mha}).
\end{itemize}

\section{Storage Considerations and Quantization} \label{sec:quantization}

In a real-world settings it may be necessary to store extracted features for later use, such as when new tasks are introduced that require ``backfilling" classifications for older content~\cite{shen2020towards}.
Storage costs quickly become impractical when pooling over multiple hidden layers and positions (cf.~Section~\ref{sec:pooling}).
For example, some of the methods that we experiment with require using the features from every layer and position in the encoder.
For RoBERTa$_\textsc{large}$, with 24 layers and 1024 dimension representations, a 50 token input would thus emit \texttt{50*24*1024} half-precision floating point numbers and require 2.3MB of storage!

We consider quantization methods, described below, for reducing the storage costs associated with extracted features.
We will show in Section~\ref{sec:res} that extracted features are surprisingly robust: they show little degredation in end-task accuracy even with binary quantization.

With quantization, we replace floating point numbers with alternative representation formats that have reduced bit width.
For example, recent work has shown that the BERT model weights and activations can be quantized down to 8-bit integers with minimal affect on downstream task accuracy~\cite{zafrir2019q8bert}.

We explore both 8-bit (uint8) and 1-bit (boolean) quantization of our extracted features.
We apply quantization prior to leave-one-task-out finetuning (see Section~\ref{sec:loto}) to simulate a real-world setting in which only the quantized features are available.
For 8-bit quantization, we use PyTorch~\cite{paszke2019pytorch} to learn scale and zero-point parameters which allow us to map floating point numbers to the range 0-255.
For 1-bit quantization, we simply apply the \texttt{sign} function to binarize each feature dimension.

Table~\ref{tab:storage} shows estimated storage costs for various pooling methods before and after quantization.

\insertStorageTable{}

\section{Results and Discussion} \label{sec:res}

\insertMainTable

Table~\ref{tab:all} presents our main results for the 14 tasks introduced in Section~\ref{sec:tasks}.
Detailed results of all tasks are included in Table~\ref{tab:extended} in Appendix.

\subsection{Baselines} \label{sec:res:baseline}

Table~\ref{tab:all} (a) shows results for models fine-tuned end-to-end on a single task.
This approach yields the best end-task accuracy but has the highest inference costs (see discussion in Section~\ref{sec:single_task}).

We observe that the distilled RoBERTa model (\texttt{DistilRoBERTa}) achieves competitive accuracy across many tasks with only 1/4 as many parameters and requiring only 1/7 of the computation of the full RoBERTa model.
Multi-task pre-training (see Section~\ref{sec:loto}) prior to single-task finetuning improves results with an average gain of +0.2\%.
This is consistent with recent work~\cite{liu2019mtdnn,wang2019can}, but somewhat at odds with the findings of~\citet{raffel2019exploring}, who report slightly worse performance with multi-task pre-training.
It remains an open question under what conditions multi-task pre-training improves end task accuracy for single-task models.

\subsection{Feature extraction and pooling} \label{sec:res:pooling}

\subsubsection{Without multi-task pre-training}

Table~\ref{tab:all} (b) shows results for single-task models where the encoder is frozen and only the classification head is fine-tuned.
We can use these results to compare to the pooling approaches described in Section~\ref{sec:pooling} that include an intermediate multi-task pre-training step.

We first observe that freezing the pre-trained RoBERTa model and extracting features from the last layer's \texttt{CLS} token performs quite poorly, with a 15\% drop in accuracy compared to the end-to-end fine-tuned version (90.5\% $\rightarrow$ 75.5\%).
This is not too surprising, since the \texttt{CLS} token is not heavily used in the RoBERTa pre-training process~\cite{liu2019roberta}.\footnote{Unlike BERT~\cite{devlin2018bert}, RoBERTa does not pre-train with a Next Sentence Prediction (NSP) objective, thus the \texttt{CLS} token is mostly unused.}
If we instead average the features across all positions in the last layer, we see slightly higher accuracy compared to using the \texttt{CLS} token alone (77.7\% vs.~75.5\%), while our multi-head attention (MHA) pooling further improves accuracy to 83.3\%, confirming the importance of task-specific position-wise pooling.

We next consider different layer-wise pooling strategies, still using the MHA position-wise pooling.
Taking a simple average over the top 16 layers improves accuracy by +2.2\% compared to using just the last layer (85.5\% vs.~83.3\%).
If we instead learn a task-specific weighted combination of layers, similar to \citet{peters2019tune}, we gain an additional +0.1\% compared to using a simple average.
However, using a task-specific combination of layers introduces significant storage costs (see Table~\ref{tab:storage}), thus we focus on the \textsc{layer-avg} pooling approach in the rest of our experiments.

\subsubsection{With leave-one-task-out multi-task pre-training}

Table~\ref{tab:all} (c) presents results for various pooling approaches after multi-task pre-training (Section~\ref{sec:loto}), in which the encoder is fine-tuned on $k-1$ tasks prior to being frozen.

In this setting, we observe that the last layer's \texttt{CLS} token now encodes quite a lot of general task information, achieving a higher average accuracy than any of the frozen encoders that did not have leave-one-task-out multi-task pre-training (85.9\% vs.~85.6\%).
As before, our multi-head attention (MHA) position-wise pooling strategy performs best, outperforming the \texttt{CLS} approach by +0.9\% and the \textsc{position-avg} strategy by +0.8\%.
Layer-wise pooling across multiple layers provides an additional 1.6-1.7\% gain.

\subsection{Quantization} \label{sec:res:storage}

Table~\ref{tab:all} (c) shows the effect of quantization on task accuracy.
We quantize extracted features after leave-one-task-out multi-task pre-training and use \textsc{layer-avg / mha} pooling, which offers the best balance between storage efficiency and accuracy.
In early experiments, we considered whether to quantize before or after layer-wise pooling and found that quantization before layer-wise pooling was slightly better for 1-bit quantization and had no impact on 8-bit quantization.

We can see that with 8-bit quantization, we don't have any performance loss.
Most surprisingly, with our 1-bit quantization method of simply applying the \texttt{sign} function on the extracted features, the average accuracy drops by only 0.4\%, while reducing the storage cost by a factor of 16 (to just 1024 bits per token) and still outperforming distillation-based methods (88.0\% vs.~87.1\%).

\subsection{Generalization} \label{sec:res:generalization}

So far we have considered encoder models evaluated in a \emph{leave-one-task-out} setting, which allows us to evaluate generalization to unseen tasks.
Another setting of potential interest is the case where an entire \emph{task type} (see Section~\ref{sec:tasks}) is held out during multi-task pre-training, e.g., an encoder may be pre-trained over all non-NLI tasks and then frozen and evaluated on NLI tasks.
We present these results in the fourth section of Table~\ref{tab:all} (d).
We observe that performance drops considerably from the corresponding leave-one-task-out setting (Table~\ref{tab:all} (c)), with average accuracy decreasing from 88.4\% to 86.6\%.
Notably, accuracy on NLI tasks decreases the most from 87.8\% to 81.3\%, consistent with past work showing significant positive transfer effects between NLI tasks~\cite{phang2018stilts}.
Thus, it seems important to pre-train the encoder over a variety of task types to maximize generalization on new tasks.

Another setting of interest is the case where the encoder is pre-trained over \emph{all $k$ tasks}, then frozen and evaluated directly on each task without any additional finetuning.
While this setting does not measure generalization to unseen tasks, it may be suitable in settings where the set of tasks is known in advance.
We present these results in the final section of Table~\ref{tab:all} (e).
We observe that this system performs 3.4\% better than a comparable system pre-trained over $k-1$ tasks and evaluated on an unseen task (89.3 vs.~85.9).

\subsection{Computational cost during inference} \label{sec:res:flops}

Table~\ref{tab:all} reports cumulative inference cost (over 14 tasks) for each method.
Single-task finetuning is the least computationally efficient approach, although it achieves the highest average accuracy (90.7\%).
Approaches based on frozen encoders reduce FLOPs by an order of magnitude, but have lower end task accuracy and extracted features may consume non-negligible storage (cf.~Table~\ref{tab:storage}).

\insertFLOPsVSNTasksFigure{}

In Figure~\ref{fig:flops_vs_ntasks} we show the number of FLOPs required for inference as a function of the number of tasks performed on the same text.
While single-task finetuning of the full model is never efficient, distilled models are in fact more efficient for systems with 7 or fewer tasks.
On the other hand, frozen encoder approaches become significantly more efficient in systems with more than 7 tasks.

\section{Conclusion} \label{sec:conclusion}

We study ways to make large-scale pre-trained models usable in practical settings. We show that when several tasks need to be performed on a single piece of text, the computation can be effectively amortized reducing the amount of computation per task. Compared to distillation approaches, the shared computation method achieves higher accuracy and the total computational cost becomes smaller after about 7 tasks. Further, we show that the shared features can be quantized with very little loss in accuracy, which means that the intermediate computation can be stored for later use. In total, the techniques that we present provide the best alternative for running large-scale pre-trained models in practical applications when multiple predictions are made on the same piece of text.

\bibliography{textray,anthology}
\bibliographystyle{acl_natbib}

\newpage

\appendix

\section{Detail Results} \label{sec:detailresults}

\insertExtendedTable{} 

The results for all setups over 14 tasks can be found in Table~\ref{tab:extended}.

\end{document}